\documentclass[preprint,12pt,number]{elsarticle}
\usepackage[utf8]{inputenc}
\usepackage[T1]{fontenc}
\usepackage{multirow}
\usepackage{subcaption}
\usepackage{mathtools}
\usepackage{graphicx}
\usepackage{float}
\usepackage{url}
\usepackage[T1]{fontenc}
\usepackage[linesnumbered,ruled,vlined]{algorithm2e}

\usepackage{amssymb}

\journal{Robotics and Autonomous Systems}

\begin{document}

\begin{frontmatter}

\title{Dynamic Q-planning for Online UAV Path Planning in Unknown and Complex Environments}

\author[1]{Lidia G. S. Rocha}

\ead{lidiarocha@ieee.org}

\author[2]{Kenny A. Q. Caldas}
\ead{kennycaldas@usp.br}

\author[2]{Marco H. Terra}
\ead{terra@sc.usp.br}

\author[3]{Fabio Ramos}
\ead{fabio.ramos@sydney.edu.au}

\author[1]{Kelen C. Teixeira Vivaldini}
\ead{vivaldini@ufscar.br}

\affiliation[1]{organization={Federal University of São Carlos, Department of Computer},
    addressline={Rod. Washington Luis, Km 235}, 
    city={São Carlos},
    postcode={13565-905}, 
    state={São Paulo},
    country={Brazil}}

\affiliation[2]{organization={São Carlos School of Engineering, University of São Paulo, Department of Electrical and Computer Engineering},
            addressline={Av. Trabalhador São-Carlense, 400}, 
            city={São Carlos},
            postcode={13566-590}, 
            state={São Paulo},
            country={Brazil}}

\affiliation[3]{organization={University of Sydney},
            addressline={G01, City Rd, Darlington, New South Wales 2006, AU}, 
            city={Sydney},
            state={New South Wales},
            country={Australia}}


\begin{abstract}
Unmanned Aerial Vehicles need an online path planning capability to move in high-risk missions in unknown and complex environments to complete them safely. However, many algorithms reported in the literature may not return reliable trajectories to solve online problems in these scenarios. The Q-Learning algorithm, a Reinforcement Learning Technique, can generate trajectories in real-time and has demonstrated fast and reliable results. This technique, however, has the disadvantage of defining the iteration number. If this value is not well defined, it will take a long time or not return an optimal trajectory. Therefore, we propose a method to dynamically choose the number of iterations to obtain the best performance of Q-Learning. The proposed method is compared to the Q-Learning algorithm with a fixed number of iterations, A*, Rapid-Exploring Random Tree, and Particle Swarm Optimization. As a result, the proposed Q-learning algorithm demonstrates the efficacy and reliability of online path planning with a dynamic number of iterations to carry out online missions in unknown and complex environments. 
\end{abstract}

\begin{highlights}

\item Utilizing Reinforcement Learning for Unmanned Aerial Vehicles to navigate through unknown and unstructured environments;
\item Introducing a dynamic iteration selection method for Q-Learning to optimize its performance in real-time path planning;
\item Comprehensive comparison with various techniques conducted through software-in-loop simulations.
\end{highlights}


\end{frontmatter}

\section{Introduction}
\label{intro}

In the last years, Unmanned Aerial Vehicles (UAVs) have gained space in several areas such as search and rescue, environmental monitoring, and surveillance ~\cite{vrba2022autonomous,bailon2022real,gyagenda2022review}. Path planning algorithms are required to perform these tasks online in unstructured and unknown environments to ensure the UAV will reach its goal safely. These techniques can analyze the environment and find trajectories to move a robot from start to goal. The main difference among these algorithms is the time to find the shortest trajectory and how optimized the trajectory is, considering path length, mission time, distance to the obstacle, curve smoothness, and battery costs.

Path planning algorithms are classified into exact classical, sampling-based classical, meta-heuristic, and machine learning techniques. When the environment is complex and unstructured, the exact classical techniques take a long time to return a trajectory~\cite{aggarwal2020path}. Classical sampling-based techniques do not return the best path, as they are based on randomness, and sometimes fail to find a path in very complex environments~\cite{aggarwal2020path}. The meta-heuristic techniques try to find the best path in an optimal time, but the complexity of the environment can make it challenging to find a path free of collisions. Besides that, their techniques present low completeness~\cite{rocha2020comparison}.

Machine learning has advanced significantly recently. These methods can be subdivided into supervised, unsupervised, and Reinforcement Learning techniques (RL)~\cite{ma2018saliency,yan2020fixed,dooraki2021innovative}.

Path planning often relies on reinforcement learning (RL), a widely used machine learning technique. RL works by learning a policy function that maps states to actions. The objective is for an agent, following this learned policy, to maximize the cumulative rewards along a sequence of actions. Then, the agent learns how to achieve the goal of completing uncertain and complex tasks. The training is a trial and error system to solve the problem, based on rewards and penalties for the actions to be performed~\cite{lison2015introduction}. They present low computational cost~\cite{song2021autonomous}, an optimal path in a short time~\cite{chen2017decentralized}, and automate much decision-making due to the control tasks in safety-critical domains~\cite{kahn2017uncertainty}. Also, the agent learns how to act in an unknown environment by repetitive iteration with that particular environment~\cite{sutton2018reinforcement}, finding an optimal strategy for interaction with it~\cite{sichkar2019reinforcement}. The main advantage of this approach is that it does not require accurate environmental models and can be used in various scenarios with little information~\cite{chen2020reinforcement}. Thus, the RL techniques generate trajectories in less time without much computational cost, even in complex environments. 

In an RL framework, an optimal trajectory can be found from agents that ought to take actions in an environment to maximize the cumulative reward. These rewards are received by movements that can be useful (a positive reward) or unhelpful, like hitting a wall (a punishment, negative reward)~\cite{lison2015introduction}.  

The RL algorithms can be classified based on their learning methods~\cite{arulkumaran2017deep,qu2020novel} as policy-based or value-based, on-policy or off-policy, and model-based or model-free:
\begin{itemize}
    \item \textbf{Policy-based:} a representation of policy is created, maintained during learning;
    \item \textbf{Value-based:} methods do not explicitly save the policy, only the value function;
    \item \textbf{On-policy:} try to improve the policy that is used to make decisions, learning from took actions from the current policy; 
    \item \textbf{Off-policy:} try to improve a different policy from that used to generate the data by learning from different actions, such as random actions; 
    \item \textbf{Model-based:} has access to the environment model to decide the best decision to make without actually taking action;
    \item \textbf{Model-free:} it does not have a model of the environment and accumulates some experience, then it gets the ideal policy.
\end{itemize}

When navigating in uncertain, the characteristics of a model-free approach with minimal computation (\(O(1)\)) and reduced space complexity (\(O(\textit{States x Actions})\)), take precedence. The absence of a pre-defined model endows the algorithm with the capability to dynamically adapt to the nuances of the environment, a crucial attribute in scenarios where the complete model is either unknown or subjected to frequent changes. The consequential gains in efficiency, stemming from diminished computational demands, become particularly significant in resource-constrained situations, facilitating swift decision-making and responsive path planning. Further enhancing its utility is the off-policy nature of the method, marked by continual exploration to assess and refine decision-making policies, proving invaluable in navigating uncertain terrains. The algorithm's adeptness to both exploration and exploitation throughout the entire environment becomes indispensable for robust path planning in complex landscapes, ensuring adaptability to unforeseen obstacles and changes~\cite{sutton2000policy}.

Moreover, Q-learning distinguishes itself as an exemplary algorithm for path planning in challenging environments. Its model-free paradigm, coupled with minimal computational requirements and perpetual exploration, positions it as an adept choice for seamlessly adapting to dynamic and intricate terrains. The algorithm's proficiency in efficiently exploiting the entire environment through value-based decision-making further amplifies its efficacy in discerning optimal paths amidst uncertainty~\cite{sutton2000policy}.

In contrast to alternative policy gradient methods, Q-Learning offers distinct advantages, excelling in simplicity, speed, and guaranteed convergence. It efficiently learns policies, especially for trajectory generation in UAVs, within short time frames. Notably adaptable, Q-Learning navigates the exploration-exploitation trade-off adeptly, simplifying complexities compared to traditional techniques~\cite{waskow2010learning, song2021autonomous}.

Compared to popular reinforcement learning methodologies like Deep Q-Network (DQN) and Double Deep Q-Network (DDQN), Q-learning is the preferred choice for path planning in unknown scenarios. DQN and DDQN, relying on deep neural networks, struggle with uncertainties in unfamiliar environments. 
Q-learning ensures convergence, ensuring stable learning outcomes compared to the complex architectures of DQN and DDQN. In complex and unknown environments, Q-learning distinguishes itself as a straightforward, adaptive, and efficient algorithm~\cite{levine2020offline}.

However, the main disadvantage of Q-Learning is its dependency on the number of iterations in training. The higher this number, the longer it takes. The value of the iteration number is usually determined empirically. Therefore, in most cases, the best results are not reached and take longer than necessary, which is unfeasible for online path planning~\cite{sichkar2019reinforcement,kim2017obstacle,kulkarni2020uav}. 

This study proposes an online path planning for UAVs based on Q-Learning using a dynamic number of iterations to carry out tasks in unknown and complex environments. Overall, our contributions are mainly in three aspects:

\begin{itemize}
    \item A Reinforcement Learning algorithm for UAV to carry out real-time tasks in unknown and complex environments;
    \item Statistically analyze from a path planning perspective the memory, CPU, path length, and time to a generated path among environments and a different number of iterations;
    \item Introduces a methodology to determine the optimal number of iterations for Q-Learning in specific environments.
\end{itemize}

The proposed method was evaluated through extensive experiments in different environments, demonstrating its efficacy and reliability for online path planning. 
The Gazebo simulator was employed to build both indoor and outdoor environments, and the tasks were done considering unknown and complex scenarios. We conducted a comparative analysis of the Q-Learning algorithm with A*, RRT (Rapid-Exploring Random Tree), and PSO (Particle Swarm Optimization), aiming to demonstrate the advantages and disadvantages of each technique.

The algorithm was tested and proven effective in both 2D indoor and outdoor environments. It is capable of delivering reliable outcomes even in situations where three-dimensional data is not available. These environments represent tangible mission scenarios where the algorithm's performance has been validated. In addition, Q-Learning uses the Q-Table to find the best path to add a third dimension without changing the algorithm's operation in 2D~\cite{wang2017autonomous,james20163d}. Therefore, the results found in 2D are similar to 3D simulations.  

This paper follows this structure. Section~\ref{sec:relatedworks} introduces related work for path planning in unknown and complex environments and path planning based on Reinforcement Learning. Section~\ref{sec:preliminaries} provides the main concepts and mathematical background used in the development of this work. 
The Q-Learning path planning approach we are proposing is presented in Section~\ref{sec:methodology}. Section~\ref{sec:simulations} shows experiments and evaluations of path planning, and in Section~\ref{sec:real} we describe how this approach can be beneficial for missions in real environments. Finally, we conclude our work and propose some directions for future work in Section~\ref{sec:conclusion}.

\section{Related Works}
\label{sec:relatedworks}
Path planning is an area of study in mobile robotics that has significantly increased in recent years, mainly with the increase of commercial drones~\cite{poikonen2017vehicle}. One of the challenges in path planning is generating an online trajectory for an unknown and complex environment~\cite{aggarwal2020path}. 

Most of the existing RL algorithms is applied to games with few movement options available. In \cite{wu2019uav}, it is proved that reinforcement learning can be successfully applied to solve problems with four movement directions. The authors also show how the algorithm can be used for UAV missions, such as oil and gas field inspection and search and rescue of injured people after a disaster.

Some Reinforcement Learning techniques, such as Q-Learning, are based on the Markov Decision Process (MDP), a mathematical framework for decision-making where the responses can be partly random and partly under the control of a decision-maker~\cite{puterman2014markov}. The algorithm does not need to know the whole environment in which it is employed to return a response~\cite{hegazy2016comparitive}, but most of the studies about path planning for UAVs tend to use previous known maps~\cite{wang2017autonomous,wu2019uav}.. 

In \cite{wang2019autonomous}, a method based on deep reinforcement learning (DRL) is proposed to solve navigation tasks in complex  large-scale environments. The problem is formulated as a Partially Observable Markov decision process (POMDP) and solved by an online DRL algorithm based on two strictly proved policy gradient theorems within the actor-critic framework. 

The authors of~\cite{singla2019memory} use a DRL-based method for UAVs to move around in unknown and unstructured indoor environments. This method's main idea is the concept of partial observability and how UAVs can retain relevant information about the structure of the environment to make better future navigation decisions. However, the authors only test the method indoors, which is usually more straightforward.

Researchers also use Reinforcement Learning to optimize other techniques, especially meta-heuristics~\cite{qu2020novel}. Suppose the Reinforcement Learning algorithm is well-optimized and uses the correct parameters. In that case, it is possible to achieve results as good as, or even better, the results found when the technique was used to optimize others. For example, in~\cite{kim2017obstacle} performed an analysis of the number of iterations of the Q-Learning algorithm, considering only the execution time.

Reinforcement Learning techniques play a pivotal role in overcoming obstacles, particularly in drone navigation scenarios, where pre-training and online adaptability are crucial~\cite{yang2019real,charlton2019fast}. For instance, in~\cite{kim2017obstacle}, an RL algorithm is employed for path planning, showcasing path generation times ranging from 15 to 52 seconds within a 20x20-meter environment, contingent on the number of iterations.

Among RL algorithms, Q-learning stands out for its model-free nature, excelling across diverse problem domains without necessitating prior knowledge of system dynamics. Its simplicity and effective exploration through an epsilon-greedy strategy enhance accessibility and applicability. Notably, off-policy learning ensures stability, enabling updates while following different policies~\cite{jang2019q}. Q-learning's convergence properties and temporal difference learning further contribute to its adaptability to unknown environments~\cite{clifton2020q}.

However, Q-learning faces challenges, notably the need to define iteration numbers, potentially resulting in suboptimal paths and extended computation times. Recognizing this limitation, researchers propose innovative approaches to enhance Q-learning's efficiency and reliability~\cite{liu2021policy}.

In~\cite{qu2020novelDeep}, the authors address Q-learning's slow convergence by integrating it with Grey Wolf Optimizer (GWO) for 3D UAV trajectory planning. This novel approach treats trajectory planning as an optimization problem, incorporating operations like exploration and intensification. Reinforcement learning controls each operation's performance, reducing the need for manual parameter definition.

Another optimization for Q-learning is presented in~\cite{LOW2019143}, where the authors introduce partially guided Q-learning. Given Q-learning's slow convergence, they integrate the flower pollination algorithm (FPA) to enhance initialization. Experimental evaluations in challenging environments demonstrate accelerated convergence when Q-values are appropriately initialized using FPA.

This work proposes an RL-based method for generating online trajectories in unknown and complex environments, both indoor and outdoor. Tests involve UAV mapping in unknown spaces and a comprehensive study on iteration numbers for RL techniques. Results show the feasibility of RL-based online path planning for UAVs, producing trajectories in just 0.08 seconds.

\section{Preliminaries} 
\label{sec:preliminaries}
In this section, we present the preliminary information for the work. We aim to generate the best path according to path length, computational time, memory, CPU, and completeness. 

\subsection{Markov Decision Process}

In a Reinforcement Learning task, the agent is required to interact with the environment to learn how to choose an action $a \in A$, where $A$ is the set of actions available, that will result in the best output based on rewards or penalties, given by the function $R$. The main objective is to find the sequence of states $s \in S$, where $S$ is the set of possible states and action pairs that will achieve the maximum reward over time. In other words, find the better policy $\pi$ defined as

\begin{align}
    \pi(a|s) = P\left[ A_t = a | S_t = s \right],
\end{align}

where $t$ is the time at any step. This problem can be formulated as a Markov Decision Process \cite{sutton2018reinforcement}. 

MDP is a mathematical definition to model the sequential decision-making task, which is defined by a tuple \textit{M = (S, A, P, R, $\gamma$)}, where:

\begin{itemize}
    \item \textbf{State space ($S$)}: The possible states where the agent can be located in the environment.
    \item \textbf{Action space ($A$)}: The actions available such as moving forward, right, left, and backward.
    \item \textbf{State-transition model ($P$)}: The probability for agents to realize each action in the state.
    \item \textbf{Reward function ($R$)}: A function that returns the reward, positive or negative, for each action.
    \item $\gamma$ determines the importance of future rewards. This parameter is used to balance immediate and future rewards.
\end{itemize}

We are using a discrete approach, where its main advantage is that it allows us to predict the next state and expects reward based only on the agent's current state, which is called the Markov property. Also, make it possible to explore the actuator's full potential at each point in time~\cite{song2021autonomous}.

\subsection{Value Functions}

The criterion to evaluate the better policy for the agent is to assign a value for each state $s \in S$. Thus, it is possible to estimate how good an action will impact in terms of the future rewards when the agent starts its progress from $s$. Using the MDP formulation, we can define a value function for the state $s$ as

\begin{align}
    v_{\pi}(s) = \mathbb{E}_{\pi} \left[ \sum_{k=0}^{\infty} \gamma^k R_{t+k+1} | S_t = s \right]. \label{eq:value_func}
\end{align}

Similarly, we can define a value for each action $a \in A$ when in a given state $s$ under a policy $\pi$, called action-value function or Q-function ($q_{\pi}(s,a)$), expressed as 

\begin{align}
    q_{\pi}(s,a) = \mathbb{E}_{\pi} \left[ \sum_{k=0}^{\infty} \gamma^k R_{t+k+1} | S_t = s, A_t = a \right]. \label{eq:action_func}
\end{align}

Based on the recursive properties of Equations \eqref{eq:value_func} and \eqref{eq:action_func} through RL and dynamic programming, the relationship between state and action value functions is given by

\begin{align}
    v_{\pi}(s) = \sum_{a \in A}\pi(a|s)q_{\pi}(s,a),
\end{align}

which defines a Bellman equation. Thus, it is possible to establish a condition between the values of $s$ and its possible future states.

Since we are interested in finding the policy that will give us the highest reward over time, we define the Bellman optimal equation around the action-value function provided by

\begin{align}
    q^{*}(s,a) = \mathbb{E}\left[ R_{t+1} + \gamma v^{*}(S_{t+1})| S_t =s, A_t, a \right]. \label{eq:opt_bellman}
\end{align}

\subsection{Q-Learning}
\label{parameters}

The challenge of applying the Bellman optimal equation defined in Equation \eqref{eq:opt_bellman} is that it is required to have a model of the environment, the state probability distributions, and its rewards. Thus, a framework called Temporal-Difference (TD) Learning is used to reduce the task's complexity. TD algorithms learn directly from trial and error experience by reevaluating their prediction after taking a step. Among these methods, Q-Learning is one of the most efficient algorithms~\cite{Jang2019}.

Q-Learning is an off-policy algorithm. Therefore, the agent learns the best way to reach the goal based on the Q-Table, which calculates each state's maximum expected future rewards. The Q-Table is initialized with random values to have any significant value during the initial phases of the exploration and is updated based on the \textit{one-step} Q-Learning as follows:

\begin{equation}
    Q(S_t, A_t) \leftarrow Q(S_t, A_t) + \alpha \left[R_{t+1} + \gamma  \max_{a}\left( Q(S_{t+1},a) - Q(S_t, A_t)\right)\right]
    \label{eq:bellman}
\end{equation} 
    
where $\alpha$ is the fraction in which the value of the Q-Table is updated (learning rate), that is, when the newly acquired information overrides old information. By directly approximating the Q-function to $q^{*}$, independent of the policy being used, the convergence of the learned function can be sped up.

There are parameters to be defined in the Reinforcement Learning algorithms ($\alpha$, $\gamma$, $\epsilon$, and $\epsilon_{decay}$), shown in Table~\ref{tab:parameters}.  In \cite{chen2020reinforcement}, it is carried out the study of the parameters of the Q-learning technique. We used the parameters' values that provided the best results in their work. In other words, the parameters that obtained the lowest number of episodes to cover and returned the shortest trajectory. The next action selection is based on an $\epsilon$-greedy approach with a $\epsilon$-decay parameter to balance the exploration and exploitation tradeoff. $\epsilon$ is the probability of choosing a random action. In other words, $\epsilon$ is the exploitation value. If it is a high value, the algorithm tends to set more random values; if it is a low value, it chooses the value on Q-table. This value cannot be trained, so this value can not be too high or too low. $\epsilon_{\text{decay}}$ reduces the value of epsilon over time, in a while, to favor exploration at the beginning and exploitation in the long term. Also, it should speed up the discovery of a valid policy early in the learning process. The algorithm for the action selection method is summarized in Algorithm~1. 

\begin{algorithm}
	\label{alg:act_sel}
	\caption{Pseudo-code for action selection} 
	Given $Q, S, \epsilon$ and $\epsilon_{\text{decay}}$ \\
    $n =$ uniform random number between 0 and 1; \\
    \If{$n < \epsilon$     }{
        a $\leftarrow$ random action from the action space $A$;} 
    \Else{
        a $\leftarrow \max_{A} Q(S,A)$
    }
    $\epsilon = \epsilon \cdot \epsilon_{\text{decay}}$ \\
	return a and $\epsilon$
\end{algorithm}

\begin{table}[htb]
    \centering
    \caption{Parameters used in Reinforcement Learning Algorithm.}
    \label{tab:parameters}     
    \begin{tabular}{lll}
        \hline\noalign{\smallskip}
        Parameter & Value & Meaning  \\
        \noalign{\smallskip}\hline\noalign{\smallskip}
        $\alpha$ & 0.9 & Learning rate \\
        $\gamma$ & 0.9 &  Discount factor \\
        $\epsilon$& 0.9 & Probability of choosing a  random action \\
        $\epsilon_{decay}$ & 0.9 & The decay of $\epsilon$ every step \\
        \noalign{\smallskip}\hline
    \end{tabular}
\end{table}

In this study, our focus revolves around identifying secure paths within an unfamiliar environment. Here, we define the state as the position of the UAV on a grid map representing the surroundings. The actions at our disposal encompass movements for the UAV, allowing it to progress forward, backward, ascend, descend, move left, or move right. Therefore, we assume the UAV goal is encoded as a reward function:
\begin{equation}
\label{eq:cost}
    r = 
    \begin{cases}
        \dfrac{max_{reward} \mbox{*} reached_{goal}}{time} & \textrm{if } reached_{goal}\\
        \mbox{-}1                                          & \textrm{if } collided_{obstacle} \\
        0                                                  & \textrm{otherwise}
    \end{cases}                                        
\end{equation}

The Q-Learning updated the Q-Table to define the best paths. The TD error is the difference in the values' variation of the future and current values in Q-Table, with the $\gamma$, and the reward shown in Equation~(\ref{eq:cost}). Thus, the value decreases according to the time, aiming to choose the shortest paths. Hence, the agent will search for the shortest path without collision, which is the one with greater reward.

The reward function is sparse because, in the future, we aim to carry out missions in real environments. Thus, we intend to use onboard computation and sparse matrix to save a significant amount of memory and speed up data processing. Some techniques use a dense matrix with gradient descent. However, we are using the main advantage of Q-Learning, that is, the velocity of the method, making it possible to generate real-time trajectories. 

We first use a discrete Q-Learning to get the shortest trajectory possible, as shown in~\cite{song2021autonomous}. Using discrete trajectories facilitates attaining an optimal time by enabling point-by-point optimization while incorporating the UAV's constraints~\cite{Foehn_2021}. Through the discretization of the trajectory, we can systematically evaluate the system dynamics and thrust limits at each point. This approach allows for iterative trajectory refinement, effectively minimizing the time required while ensuring compliance with the specified constraints. This method is fast because it is applied in an unknown environment. In other words, the lower complexity of the environment reduces the time required to solve the problem. 

However, for a UAV to execute the trajectory is better to use continuous trajectory due to the UAV constraints. Then, the best path is smoothed at the end of the algorithm. The smoothed path does not increase the algorithm's performance but improve the UAV motion, helping keep low jerk, constant torque, and continuous movement. Also, it makes the trajectory continuous.

\subsection{Planner}
\label{planner}
We implemented this technique on Plannie~\cite{9836102}, a path-planning framework that facilitates the use of path-planning techniques in simulated and natural environments. This framework uses a security module, which penalizes the trajectories that try to enter dangerous areas but does not prohibit them. This way, the planner will replan the trajectory to avoid the obstacles, even if the technique recommends flying in proximity to them. 

The method used to identify dangerous areas was defined by~\cite{sankararaman2018computational} and considered mainly environmental uncertainties and performance constraints. The paper~\cite{9836102} analyzed the advantages and disadvantages of each risk zone created by the security module. In this paper, we use a risk zone of 1 meter according to the realized analysis. This size of risk zone has the best trade-off between security and the amount of error in the trajectory.

\section{Methodology}
\label{sec:methodology}
This paper aims to develop an online path-planning algorithm based on Reinforcement Learning for UAVs to carry out tasks in unstructured and unknown environments. The enhancement focuses on making the number of iterations for Q-Learning dynamic. This ensures that, for all environments, the algorithm utilizes a minimal number of iterations to find a reliable path. Additionally, the cubic spline interpolation is applied to the final trajectory to make it smooth and easier for the UAV to follow~\cite{pandey2018three}.

This section outlines the approach taken for UAV navigation in an unknown environment. It covers the identification of the environment, the optimization of iteration numbers for Q-Learning, and the transformation of a discrete trajectory into a continuous one. These steps are carefully designed to ensure a smooth integration that optimally aligns with UAV constraints.

\subsection{Mapping Unknown Environment}
\label{subsec:mapping}
In missions where the UAV is unaware of the environment~\cite{cetin2016real,hayat2017multi,san2018uav}, it must identify the obstacles around it and make a global map that is constantly updated.

\begin{figure}[H]
    \centering
    \begin{subfigure}[b]{0.32\textwidth}
        \centering
        \includegraphics[width=\textwidth]{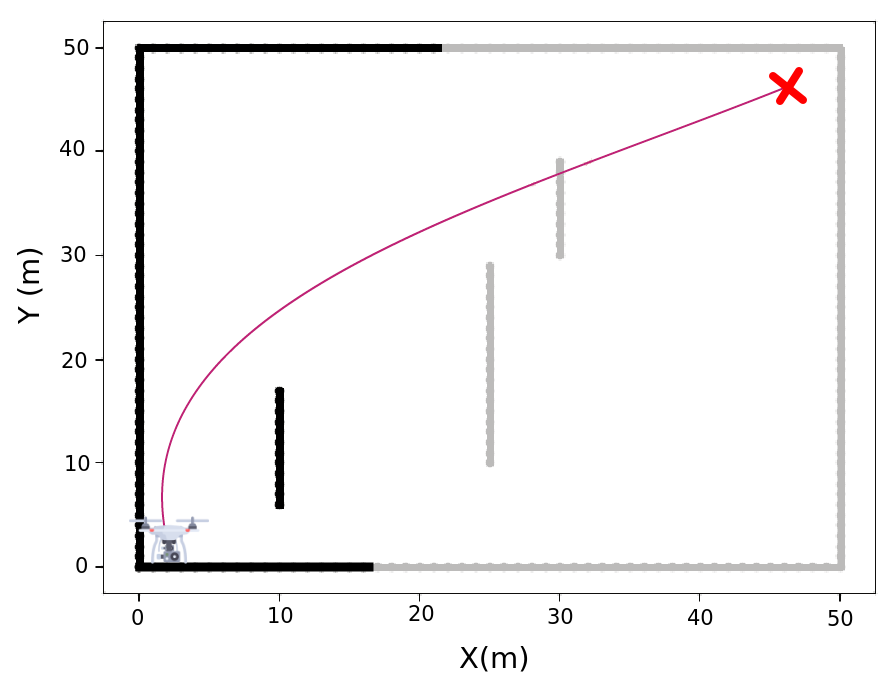}
        \caption{}
    \end{subfigure}
    \hfill
    \begin{subfigure}[b]{0.32\textwidth} 
        \centering
        \includegraphics[width=\textwidth]{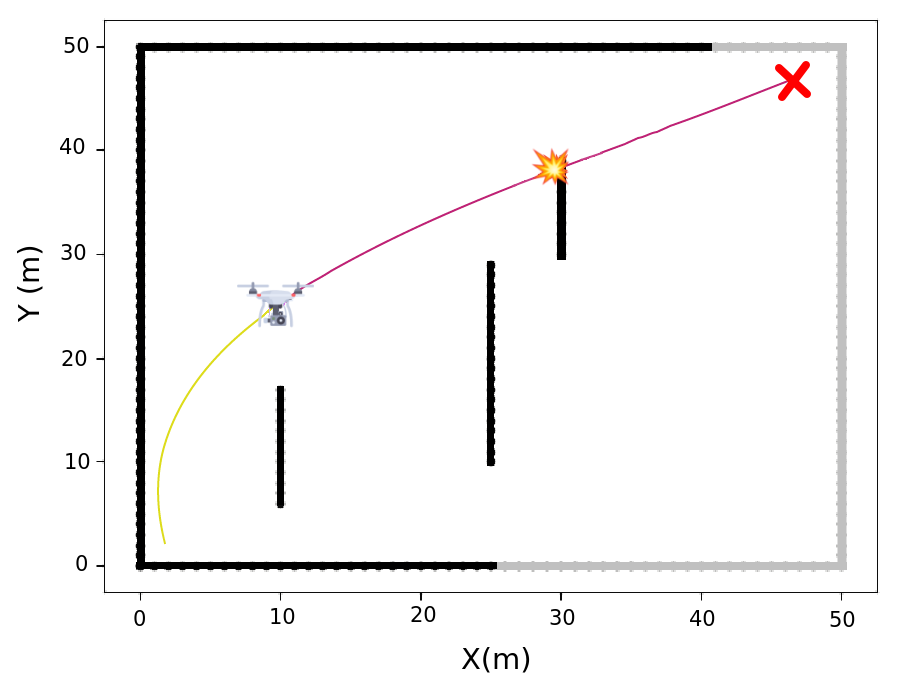}
        \caption{}
    \end{subfigure}
    \hfill
    \begin{subfigure}[b]{0.32\textwidth}
        \centering
        \includegraphics[width=\textwidth]{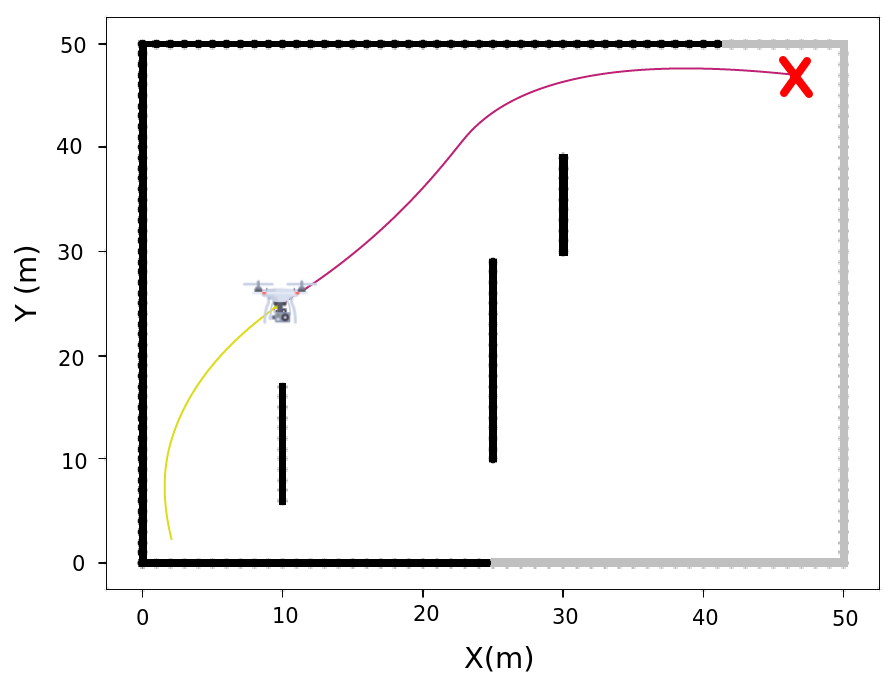}
        \caption{}
    \end{subfigure}
    \caption{The process for the UAV discovers the unknown environment: (a) Generating a trajectory in an accessible path, (b) Identifying collision along the trajectory, and (c) Replanning the trajectory to avoid the obstacle due to the collision.}
    \label{fig:replanning}       
\end{figure}

To identify obstacles, we use the open-source LIDAR-based obstacle detector package~\cite{kan2020online}. This package returns the distance measurements between the UAV and the obstacles across all sensor beams. The mapping of an environment can be seen in Figure~\ref{fig:replanning}, in which the UAV position represents the current node, the objective node is depicted as a red $X$, collision occurrences are marked by orange stars, the yellow path indicates the distance traveled, the pink trajectory represents the next steps, and the gray and black areas denote undiscovered and discovered walls, respectively.

In Figure (a) \ref{fig:replanning}, the UAV starts from node (1, 1) and can observe a partial map of the environment. 
This planning process occurs without the full knowledge of the entire surroundings. The UAV proceeds along its trajectory, it dynamically updating the map by incorporating new obstacles detected. Consequently, the UAV actively scans for potential collisions along the planned path, as illustrated in Figure~\ref{fig:replanning} (b). If a collision is detected along the trajectory, it is promptly replanned while the UAV continues to follow the adjusted path, as depicted in Figure~\ref{fig:replanning} (c).

Whenever an obstacle is detected along the trajectory, a new path is recalculated. This process involves rerunning the Q-Learning algorithm which, due to multiple iterations, can be time-consuming. However, the method introduced in this paper offers a streamlined approach, enabling the derivation of an optimal trajectory in a significantly shorter timeframe. This efficiency facilitates the feasibility of real-time flights. Notably, all iterations are grounded in the environmental model detailed in Section~\ref{parameters} and are components of the Q-Learning algorithm.

\subsection{Dynamic Iterations}
In addition to the parameters presented in Section~\ref{parameters}, defining the number of iterations for algorithm training is crucial to ensure optimal performance, especially in generating trajectories within the shortest possible time.

If the number of iterations is minimal, there is a high probability that the algorithm will not generate a trajectory or take longer than expected. It occurs because, in the training phase, insufficient information was obtained. In the test phase, the algorithm will take longer to find a collision-free trajectory, likely resulting in a considerable duration. If the number of iterations is too high, the testing phase quickly returns the shortest trajectory to follow. However, a considerable amount of time will be consumed during the training phase.

In the literature, most studies have employed a significant number of iterations, necessitating offline execution to accommodate the time required for trajectory calculations. The rewards tend to stabilize after a certain number of iterations, depending on whether the environment is indoor or outdoor. Specifically, in relatively more straightforward indoor environments, the convergence of rewards is achieved at approximately 150, 300, and 500 iterations~\cite{jiang2019path,low2019solving}. Conversely, in more complex outdoor environments, a higher number of iterations, namely around 800, 1000, and 1500, is typically required for convergence~\cite{yang2020multi,bae2019multi}.

However, the best number of iterations will not be the same for all scenarios, as it is directly proportional to the size and number of environmental obstacles. Aiming to generate the shortest trajectory in the shortest possible time, using the ideal number of iterations in Q-Learning training is necessary.

In the context of Q-Learning, the reward obtained at the end of each iteration plays a crucial role. By monitoring the stability of the reward values over consecutive iterations, one can ascertain if the algorithm has successfully optimized the trajectory. Specifically, when the reward remains constant or exhibits only slight variations from the initial to the final reward after a sufficient number of iterations, it indicates that the algorithm has effectively converged and found the optimal trajectory. The number of iterations to check if the algorithm converged is proportional to the complexity of the environment. Thus, a comprehensive complexity formula has been developed, incorporating several factors, including the size of the environment, the number of obstacles that the drone identified until the moment, the distance to the goal node, and the spatial distribution factor (SDF), as shown in Equation~\ref{eq:complexity}. This equation is based in the results of this study~\cite{a3d2022rocha}, which describes the complexity of several path planning algorithms.

\begin{equation}
\label{eq:complexity}
    complexity = (\frac{num\_obstacles}{grid\_size^2}) \times ( \frac{goal\_distance^2}{avg\_distance_{expected}} ) \times ( \frac{sdf}{max\_sdf} )
\end{equation}

The $grid\_size$ denotes the size of the discretized cells or units used to represent the environment. It is typically expressed as the length or width of a square grid cell. Determining the grid size involves measuring the length or width of a single cell in the environment grid representation.

The $num\_obstacles$ refers to the total amount of obstacles present in the environment. These obstacles can vary and encompass physical objects, walls, or other entities that must be avoided during path planning.

The $goal\_distance$ node signifies the distance between the starting point and the goal node in the environment.

The $avg\_distance_{expected}$ denotes the anticipated average distance between obstacles in the environment. It can be estimated based on prior knowledge or assumptions about the environment or derived from historical data.

The spatial distribution factor ($sdf$) captures the arrangement and distribution of obstacles within the environment. It is calculated based on the Equation~\ref{eq:spatial_distribution}.

\begin{equation}
\label{eq:spatial_distribution}
    sdf = \frac{avg\_distance_{current}}{avg\_distance_{expected}} \times grid\_size
\end{equation}

The spatial distribution factor considers the actual average distance, which is the average distance between obstacles in the environment, and relates it to the expected average distance. By multiplying this ratio with the grid size, the spatial distribution factor provides a quantitative measure of the spatial distribution pattern in the environment.

Lastly, the $sdf$ represents the highest possible value for the spatial distribution factor in a given environment. It serves as a normalization factor and can be determined based on the specific characteristics of the environment under consideration.

A comprehensive measure of environmental complexity can be obtained by incorporating all these components into the complexity formula, offering a better understanding of the challenges and requirements associated with a particular environment. This facilitates the selection of appropriate path planning algorithms and strategies, guiding the development of robust and efficient navigation systems tailored to the specific complexities of the environment. The complexity value is the same amount of iteration that needs to be monitored to check if the algorithm has converged. 

With this method, it is possible to execute numerous iterations, allowing the algorithm to continue training until it discovers the shortest trajectory, generating an efficient path in the shortest possible time. This approach ensures that the algorithm has an ample opportunity to explore and exploit the available environment, leading to a well-refined and optimal solution. By providing a sufficient number of iterations aligned with the environment's size, a balance can be achieved between training time and trajectory optimization. This strategy enables the algorithm to thoroughly navigate the environment, learn from its experiences, and refine its decision-making process, ultimately converging to an efficient trajectory.

In summary, the number of iterations in Q-Learning is essential for achieving optimal trajectory generation. By monitoring the stability of reward values and allowing several iterations commensurate with the size of the environment, assurance can be established that the algorithm converges to the shortest trajectory. This approach allows for thorough exploration, learning, and decision refinement, resulting in an efficient path that minimizes the time required to reach the desired goal.

Real-time path planning and replanning have become feasible using the presented approach, enabling UAVs to conduct flights in real-world scenarios. At the beginning of the flight, when the UAV has no prior knowledge of the environment, the Q-Learning algorithm is employed to plan the optimal path based on the available environmental information. By appropriately selecting the number of iterations, the algorithm explores and exploits the known environment, achieving maximum optimization.

As the UAV progresses through the flight, it gains additional information about the environment, mainly through identifying obstacles as discussed in Subsection~\ref{subsec:mapping}. Subsequently, a recalculation of the path becomes necessary 
due to the identification of newly discovered obstacles. The algorithm dynamically adjusts the number of iterations required to explore and exploit the environment based on the current knowledge. This adaptive approach ensures that the UAV allocates only the necessary iterations to achieve an optimal path, effectively incorporating the updated environmental information.

By iteratively updating the path planning process in response to the evolving understanding of the environment, the algorithm enables the UAV to navigate through complex scenarios while minimizing computational resources. The dynamic adjustment of the iterations allows for efficient exploration and exploitation, resulting in timely and accurate path planning and replanning. 
This capability offers a practical and effective solution for real-time path planning in continuously updated real-world environments.

Specifically, sharp turns, extending from the initial to the final control node, are substituted with new trajectories created by cubic splines.\subsubsection{Smoothing Trajectory}
\label{subsub:minimize}
To address the intricate design and operational challenges faced by UAVs and to optimize their trajectories for high performance, considerations must be made, particularly focusing on the minimization and smoothness of curves. This paper delves into the implementation of approaches aimed at generating flight trajectories that are both smooth and efficient.

The trajectories generated by Q-learning are discrete, often leading to paths characterized by abrupt curves that pose challenges for UAVs in execution. To optimize UAV navigation, it is essential to ensure that the generated paths are as smooth and straight as possible, minimizing jerk and maintaining a constant torque during flight~\cite{goel2018three}. Achieving this involves scrutinizing the trajectory to identify and eliminate unnecessary points. For instance, consider a scenario with nodes 1, 2, and 3. If there is no collision detected between nodes 1 and 3, Algorithm~2 can be applied to remove node 2 from the path. This strategic elimination of unnecessary waypoints ensures that the resulting trajectory remains as straight as possible without compromising collision avoidance with obstacles. This meticulous trajectory refinement is crucial for enhancing the feasibility and efficiency of UAV missions, particularly in environments where path smoothness is paramount. 

\begin{algorithm}
	\label{alg:refineTrajectory}
	\caption{Trajectory Refinement}
      \For{$i \gets 1$ to $|nodes| - 2$}{
        \For{$j \gets i + 2$ to $|nodes|$}{
            \If{$\text{NoCollision}(nodes[i], nodes[j])$}{
            $nodes \gets \text{RemoveNode}(nodes, i + 1)$
             }
        }
     }
\end{algorithm}



The ideal trajectory for a UAV requires smooth curves for optimal performance~\cite{pandey2018three}. In this paper, splines are employed to ensure the smoothness of trajectory curves, allowing us to calculate the best heading for each position in the trajectory and facilitating UAV movement. Optimizing curve trajectories allows for a more fluid and efficient UAV movement. This not only enhances the trajectory but also transforms the discrete trajectory generated from Q-Learning into a continuous trajectory using the Cubic Spline Interpolation Algorithm. 


Cubic Spline interpolation is a method involving third-degree polynomial segments. This algorithm divides the interval of interest into subintervals and interpolates based on cubic polynomials~\cite{mckinley1998cubic}. The cubic spline maintains continuity and smoothness at the knots, with two additional parameters at each knot point specifying the function's second derivative.

This method starts by selecting $N$ interpolation nodes $\mbox{(}x_1, y_1\mbox{)}$,  $\mbox{(}x_n, y_n\mbox{)}$, ..., $\mbox{(}x_nN, y_nN\mbox{)},$ between the start point and the objective, to determine the intervals. The spline nodes are used then to obtain the $m$ interpolation points $x_1, ..., x_m$ and $y_1, ..., y_m$, forming a continuous path when connected.

According~\cite{toraichi1987computational}, non-periodic splines, as as utilized in our approach,  
have a low complexity of $O(3n)$, which has negligible impact on the time and complexity of our online path planning.


While splines serve as efficient path-planning algorithms due to their low computational complexity, they may not be sufficient in complex environments with sharp corners, multiple obstacles, or forested terrain~\cite{liang2018geometrical}. Splines primarily aim to smooth existing curves or find smooth paths between two given nodes, making them less effective in navigating environments with diverse challenges.

To address these limitations, our approach employs cubic splines within other path-planning algorithms to enhance curve smoothing. In this configuration, the control points align with the curve nodes, ensuring sufficient maneuvering space for the UAV at optimal angles. 
Specifically, sharp turns, extending from the initial to the final control node, are substituted with new trajectories created by cubic splines. This incorporation ensures that the UAV navigates more effectively through complex environments, mitigating the challenges posed by sharp corners and obstacles, and optimizing its overall trajectory for improved performance.

\section{Simulations}
\label{sec:simulations}
This section presents the trajectory analysis of the splines used in path planning algorithms. And then, we show and analyze the experiments done indoors and outdoors. 

\subsection{Representing Trajectories with Splines}
\label{results:simple}
Before the path planning analysis was made, an analysis to understand the advantages and disadvantages of using splines in path planning algorithms, the tests were made in two known environments. The first environment has 18x15$m$ with simple obstacles, and the second environment has 100x100$m$ with more complex obstacles. The path planning algorithm chosen was A*, because it is one of the most consolidated algorithms and always returns the same trajectory, thereby decreasing the bias of the data.

\begin{figure*}
    \centering
    \begin{subfigure}[t]{0.32\textwidth}
        \centering
        \includegraphics[width=\linewidth]{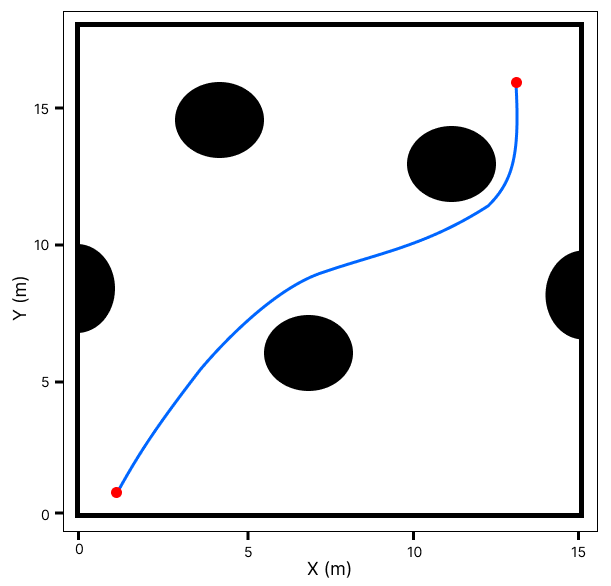}
        \caption{}
    \end{subfigure}
    \hfill
    \begin{subfigure}[t]{0.32\textwidth}
        \centering
        \includegraphics[width=\linewidth]{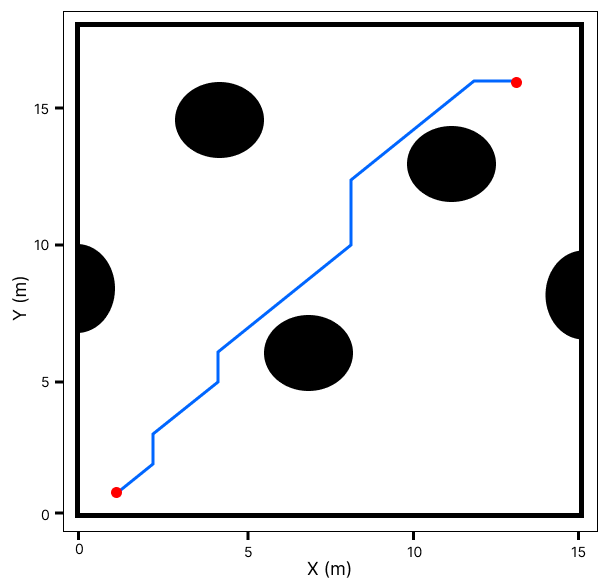}
        \caption{}
    \end{subfigure}
    \hfill
    \begin{subfigure}[t]{0.32\textwidth}
        \centering
        \includegraphics[width=\linewidth]{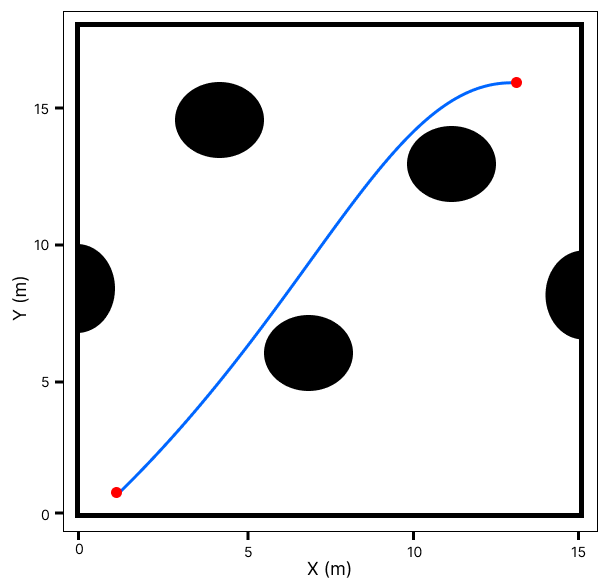}
        \caption{}
    \end{subfigure}
    \caption{Python simulations in a small and simple environment. In (a) Trajectory generated just with spline (b) Trajectory generated just with A* algorithm (c) Trajectory generated with A* algorithm with the spline.}
    \label{fig:splinesSimple}       
\end{figure*}

\begin{figure*}
    \centering
    \begin{subfigure}[t]{0.32\textwidth}
        \centering
        \includegraphics[width=\linewidth]{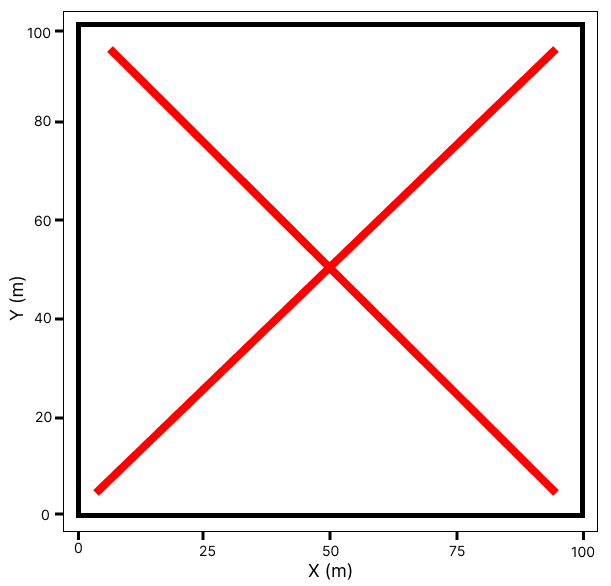}
        \caption{}
    \end{subfigure}
    \hfill
    \begin{subfigure}[t]{0.32\textwidth}
        \centering
        \includegraphics[width=\linewidth]{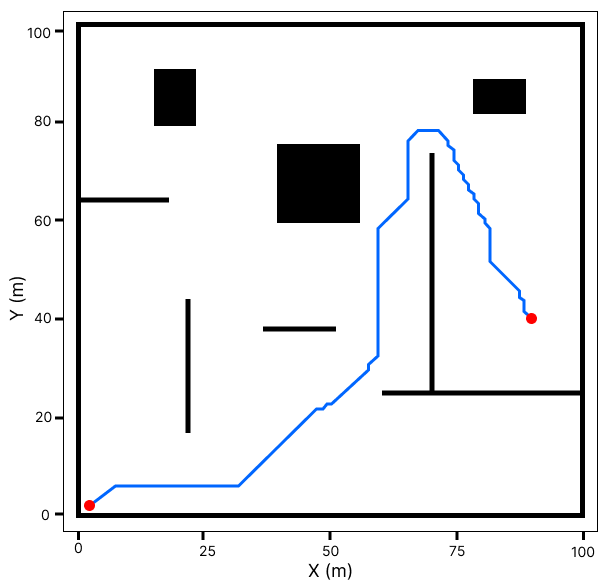}
        \caption{}
    \end{subfigure}
    \hfill
    \begin{subfigure}[t]{0.32\textwidth}
        \centering
        \includegraphics[width=\linewidth]{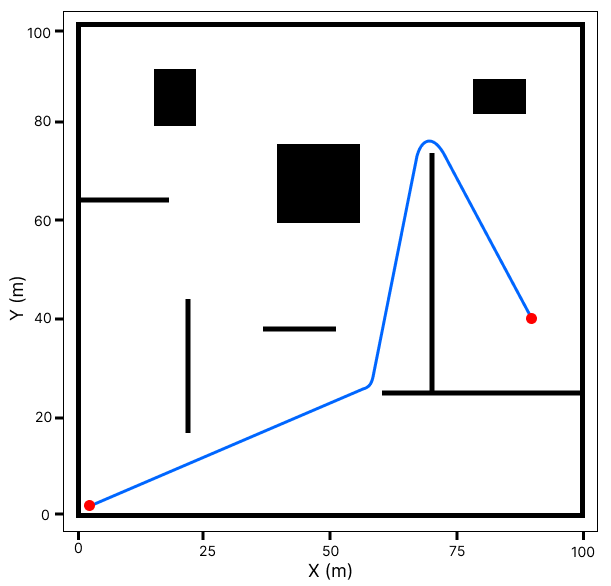}
        \caption{}
    \end{subfigure}
    \caption{Python simulations in a large and complex environment. In (a) Trajectory generated just with spline (b) Trajectory generated just with A* algorithm (c) Trajectory generated with A* algorithm with the spline.}
    \label{fig:splinesComplex}       
\end{figure*}

The simulations of small and large environments can be shown in Figures \ref{fig:splinesSimple} and \ref{fig:splinesComplex}, respectively. From a visual analysis, the trajectories generated by splines and the A* algorithm with splines are good trajectories to be followed by a UAV. However, the trajectories just with the A* algorithm, which the UAVs will have problems following due to their aerodynamic constraints. Moreover, noticeable splines can not find a reliable trajectory in large and complex environments, as was discussed in Section~\ref{subsub:minimize}.

Table~\ref{tab:splineStats} shows the previous simulations' time and path length results. We can see the use of splines with a path planning algorithm, the time to find a reliable trajectory is considerably superior to other simulations and, although the path length is just 6\% greater than others, still returns the longer trajectory in the longer time.

\begin{table}[htb]
    \centering
    \caption{Statistical Analysis according to Number of Iterations in each Environment.}
    \label{tab:splineStats}     
    \begin{tabular}{llcc}
    \hline
    Envs                   & Type           & Time (s) & Path Length (m) \\ \hline
    \multirow{3}{*}{Small} & Spline         & 2.43     & 19.96           \\
                           & A*             & 0.01     & 19.83           \\
                           & \textbf{A* with Spline} & \textbf{0.01}     & \textbf{18.88}           \\ \hline
    \multirow{3}{*}{Large} & Spline         & $\infty$   & $\infty$             \\
                           & A*             & 2.15     & 170.06          \\
                           & \textbf{A* with Spline} & \textbf{2.22}     & \textbf{157.8}           \\ \hline
    \end{tabular}
\end{table}

For small and large environments, splines with a path planning algorithm return a shorter trajectory with more time than just a path planning algorithm (3\% or 0.07s). Despite this slight trade-off, the joint utilization of A* with splines ensures not only optimal trajectory length but also adherence to UAV constraints. This comprehensive approach, balancing efficiency and compliance with operational limitations, establishes the superiority of employing splines within a path planning algorithm for generating reliable and UAV-constraint-compliant trajectories.

\subsection{Path Planning Analysis}
\label{results:analysis}
\begin{figure}
    \centering
    \begin{subfigure}[t]{0.45\columnwidth}
        \centering
        \includegraphics[width=\textwidth]{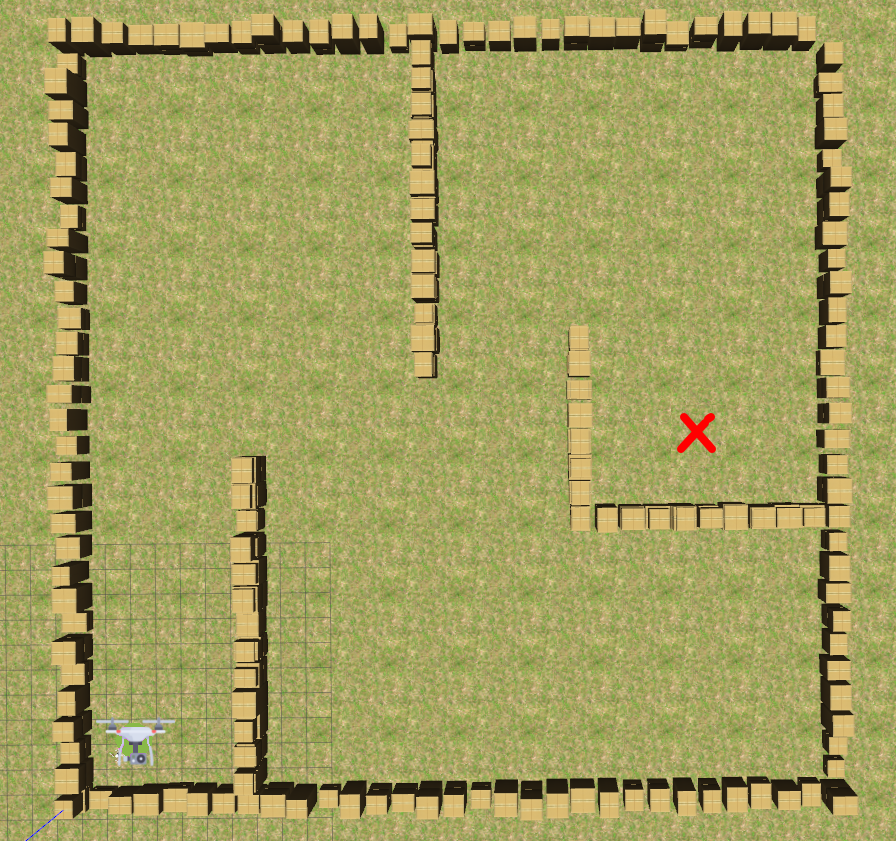}
        \caption{}
    \end{subfigure}
    \hfill
    \begin{subfigure}[t]{0.45\columnwidth}
        \centering
        \includegraphics[width=\textwidth]{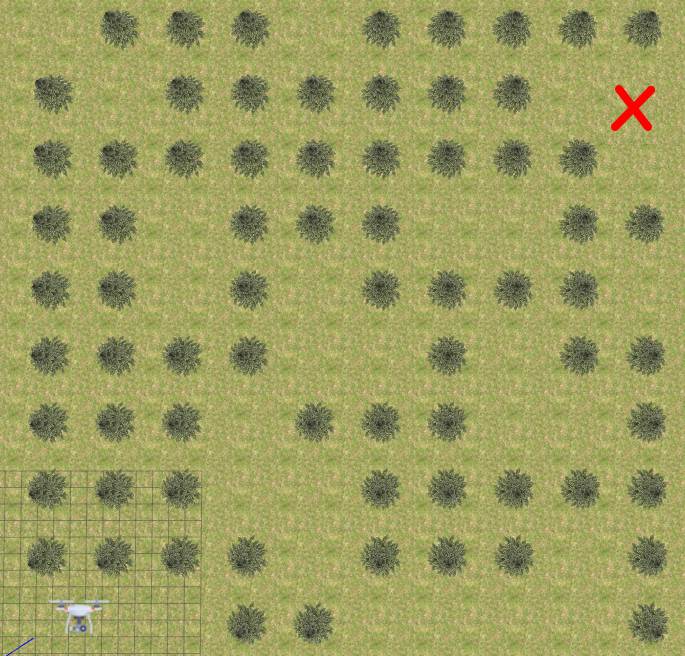}
        \caption{}
    \end{subfigure}
    
     \begin{subfigure}[t]{0.45\columnwidth}
        \centering
        \includegraphics[width=\textwidth]{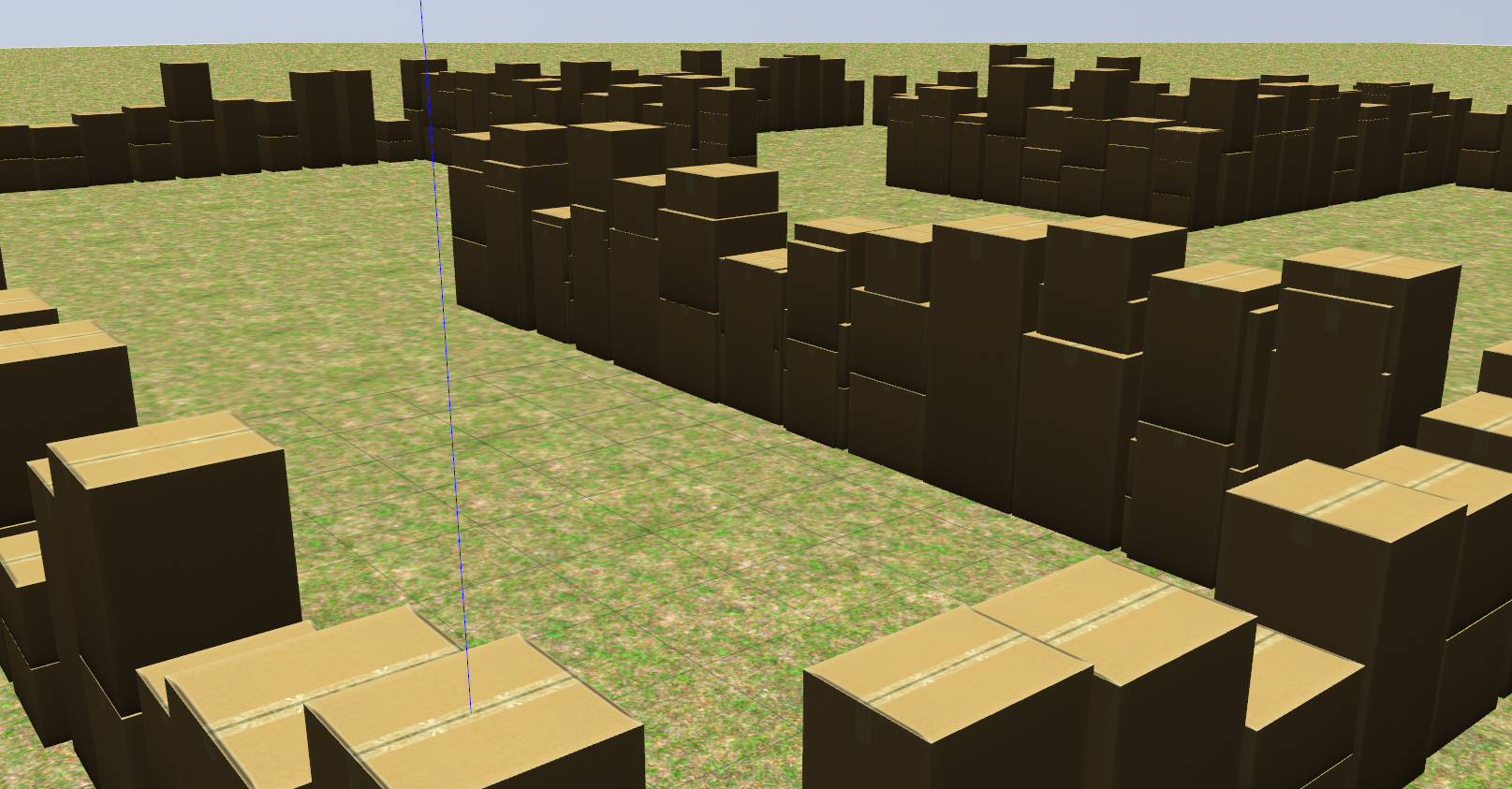}
        \caption{}
    \end{subfigure}
    \hfill
    \begin{subfigure}[t]{0.45\columnwidth}
        \centering
        \includegraphics[width=\textwidth]{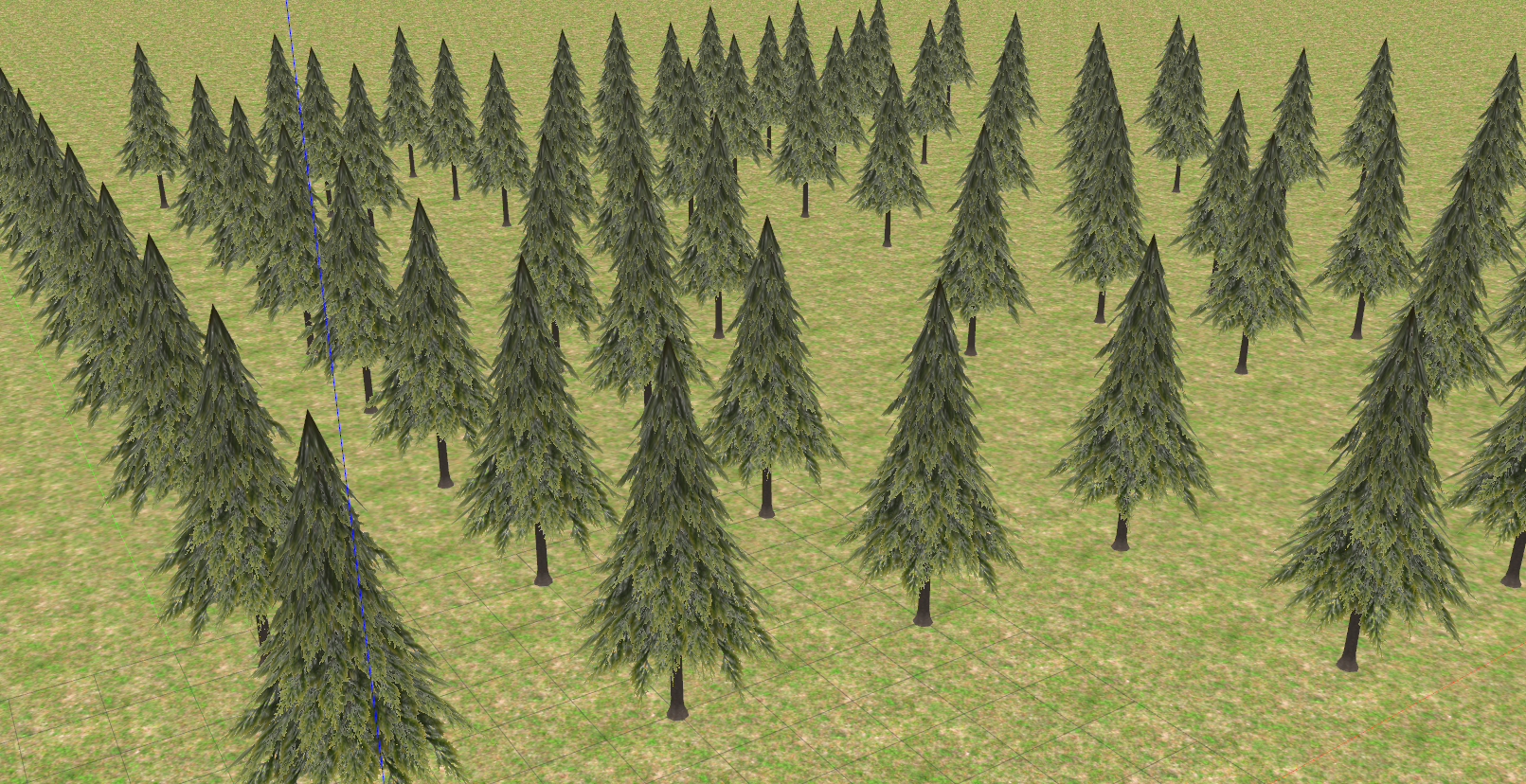}
        \caption{}
\end{subfigure}
\caption{Top and side view of the environments. The indoor environment is constructed of cardboard boxes, while the outdoor environment consists of trees. In (a) Top view of the indoor environment, (b) Top view of the outdoor environment, (c) Side view of the indoor environment, and (d) Side view of the outdoor environment.}
\label{fig:ambienteTV}       
\end{figure}

This analysis considers two different environments. The first is an indoor environment with 30x30 meters, and the second is an outdoor environment with 40x40 meters, both shown in Figure~\ref{fig:ambienteTV} with a top view. The UAV picture marks the start nodes, and red $X$ marks the end nodes. In the indoor environment, the start node is (2,~2), and the end node is (25, 14). In the outdoor environment, the start node is (2, 2), and the end node is (36, 32). Each simulation was repeated 100 times.

The environments were made using the Gazebo simulator with the Robot Operating System (ROS). Additionally, we employed the MRS system, offering a comprehensive control and estimation framework to facilitate flights within realistic simulations. Leveraging Software-in-the-loop (SITL) testing, this system allows for seamless replication in real-world experiments, as highlighted in~\cite{baca2021mrs}.

In this system, we use the UAV F-450. It is a quadcopter kit that is widespread in the market, easy to acquire and assemble, and has a low-cost~\cite{dji2015f450}. Moreover, this model support the use of an onboard computer, such as NVIDIA's Jetson, and many different sensors like cameras, IMU and Lidar for state estimation in the environment~\cite{px42020pix,pritsker1999simulation}.

In this work, the sensor used to map the environment was the RPLidar. This sensor is a 2D Lidar sensor with a range of 360$^{\circ}$ and was positioned on the top of the UAV. 


The simulations were developed using Python 3.8 on a laptop with an Intel Core i7 7700HQ processor with 16 GB of RAM and an Intel\textsuperscript{®} HD Graphics 620. The algorithms used were A*, exact classic algorithm, RRT, sampling-based classic algorithm, PSO, meta-heuristic algorithm, and Q-Learning, with various iterations. These algorithms were implemented from framework Plannie~\cite{9836102} adapting the Q-Learning algorithm to use dynamic iteration in the Q-Learning algorithm. Thus, it intends to show the difference between each technique for the scenarios. The parameters of each algorithm are shown in Table~\ref{tab:parametersOthers}. 

\begin{table}
    \centering
    \caption{Parameters for each algorithm.}
    \label{tab:parametersOthers}     
    \begin{tabular}{lll}
        \hline\noalign{\smallskip}
        Algorithm & Parameters & Value  \\
        \noalign{\smallskip}\hline\noalign{\smallskip}
        A* & Robot radius & 40 cm \\
        & Path resolution &  0.1\% \\
        \noalign{\smallskip}\hline
         & Robot radius &  40 cm \\
         & Path resolution & 0.1\% \\
        RRT  & Sample rate & 5\% \\
           & Expansion distance & 5 meters \\
            & Max iterations & 5000 \\
        \noalign{\smallskip}\hline
         & Robot radius &  40 cm \\
         & Personal learning coefficient & 1.5 \\
         PSO & Global learning coefficient & 1.5 \\
           & Inertia weight & 1 \\
            & Damping ratio & 0.98 \\
        \noalign{\smallskip}\hline
    \end{tabular}
\end{table}

\begin{figure}
    \centering
    \begin{subfigure}[t]{0.45\columnwidth}
        \centering
        \includegraphics[width=\linewidth]{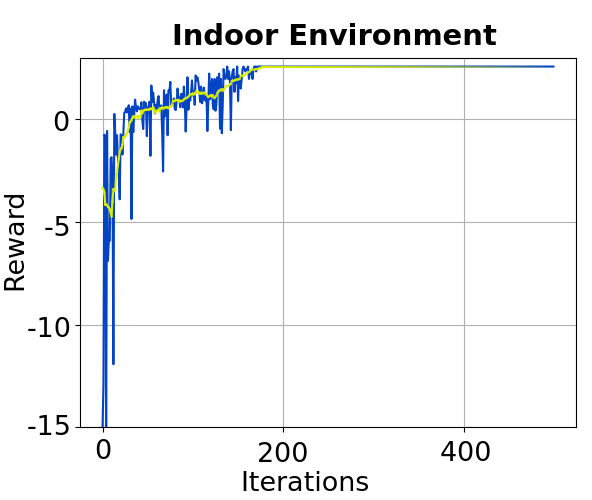}
        \caption{}
    \end{subfigure} 
    \hfill
    \begin{subfigure}[t]{0.45\columnwidth}
        \centering
        \includegraphics[width=\linewidth]{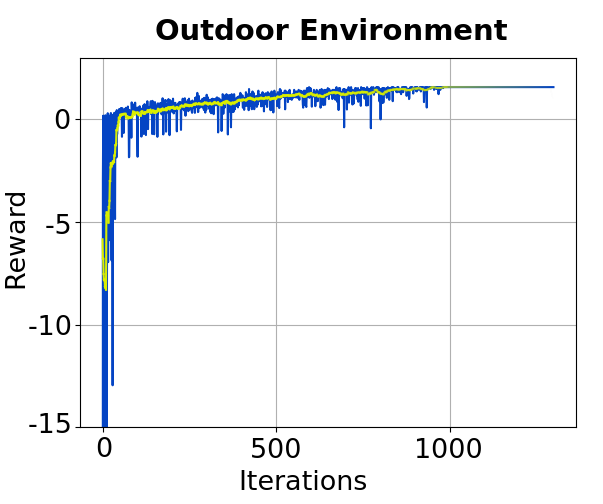}
        \caption{}
    \end{subfigure}
    \caption{Rewards of Q-Learning algorithm for each iteration. In (a) indoor environment and (b) outdoor environment.}
    \label{fig:reward}       
\end{figure}

Figure~\ref{fig:reward} shows each environment's Q-Learning reward with many iterations. The blue line presents the reward, and the yellow the moving average. The reward stabilizes at 180 iterations in the indoor and outdoor environments at 1000 iterations. Therefore, indoor simulations are performed with 150, 300, 500, and dynamic iterations. Also, there are 800, 1000, 1500, and dynamic iterations in outdoor simulations. Values below the mean, on the mean, and above the mean were chosen to show the difference when the iteration number used is unknown.

Table~\ref{tab:commands} shows the metrics, traveled distance and time, of each simulation varying in the algorithms, number of iterations, and the environment. The metrics analyzed were:
\begin{itemize}
    \item \textbf{Average distance} (Avg Distance - meters) with the standard deviation;
    \item \textbf{Average time} of the path planning algorithm considering  iteration time (Avg Time - seconds) with the standard deviation;
    \item \textbf{Best time} of the path planning algorithm considering  iteration time (seconds);
    \item \textbf{Memory} (MB);
    \item \textbf{CPU} (MHz);
    \item \textbf{Completeness} (\%).
\end{itemize}

\begin{table*}
\caption{Statistical Analysis according to Number of Iterations in each Environment.}
\label{tab:commands} 
\resizebox{\textwidth}{!}{
\begin{tabular}{llllllll}
\hline\noalign{\smallskip}
Envs & Algorithms & Avg Distance (m) & Avg Time (s) & Best Time (s) & Memory (MB) & CPU (MHz) & Completeness (\%)\\
\noalign{\smallskip}\hline\noalign{\smallskip}
\multirow{11}{*}{\rotatebox[origin=c]{90}{Indoor    }}
    &A*&\textbf{35.22$\pm$0} & \textbf{0.8$\pm$0.06} & 0.12 & \textbf{0.98$\pm$0.23} & 685.63$\pm$462.68 & 100 \\
    &RRT& 35.35$\pm$1.02 & 0.96$\pm$0.22 & 0.42 & 1.01$\pm$0.22 & 504.44$\pm$456.83 & 65 \\
    &PSO& 48.49$\pm$11.43 & 1.5$\pm$1.81 & 0.18 & 4.43$\pm$2.72 & \textbf{244.86$\pm$203.51} & 31 \\
    
    &\begin{tabular}{@{}c@{}}Q-Learning \\ 150 iterations\end{tabular}& 36.54$\pm$0.85 & 1.01$\pm$0.1 & 0.75 & 1.09$\pm$1.08 & 520.14$\pm$361.13 & 60 \\
    
    &\begin{tabular}{@{}c@{}}Q-Learning \\ 300 iterations\end{tabular}& 35.46$\pm$1.26 & 0.84$\pm$0.22 & 0.21 & 1.03$\pm$0.24 & 448.67$\pm$205.74 & 100 \\
    
    &\begin{tabular}{@{}c@{}}Q-Learning \\ 500 iterations\end{tabular}& 35.39$\pm$1.22 & 0.93$\pm$0.25 & 0.13 & \textbf{0.98$\pm$0.23} & 439.74$\pm$370.44 & 100 \\
    
    &\begin{tabular}{@{}c@{}}Q-Learning \\ Dynamic iterations\end{tabular}& \textbf{35.37$\pm$1.17} & \textbf{0.74$\pm$0.17} & \textbf{0.08} & \textbf{1.01$\pm$0.24} & \textbf{413.74$\pm$74.55} & 100 \\
    
    \noalign{\smallskip}\hline\noalign{\smallskip}
    \multirow{11}{*}{\rotatebox[origin=c]{90}{Outdoor}}
    
    &A*& \textbf{45.67$\pm$0} & 0.96$\pm$0.42 & \textbf{0.03} & 2.26$\pm$1.01 & 656.14$\pm$442.09 & 100 \\
    &RRT& 48.34$\pm$1.74 & \textbf{0.2$\pm$0.09} & 0.11 & 2.26$\pm$1.43 & 401.66$\pm$253.61 & 71 \\
    &PSO& 51.06$\pm$7.09 & 3.09$\pm$3.61 & 0.32 & 2.2$\pm$1.58 & 400.79$\pm$332.03 & 71 \\
    
    &\begin{tabular}{@{}c@{}}Q-Learning \\ 800 iterations\end{tabular}& 47.9$\pm$0.18 & 2.91$\pm$0.18 & 2.61 & 3.48$\pm$2.17 & \textbf{305.71$\pm$215.78} & 49 \\
    
    &\begin{tabular}{@{}c@{}}Q-Learning \\ 1000 iterations\end{tabular}& 47.56$\pm$1.08 & 2.77$\pm$1.3 & 1.03 & 1.13$\pm$1.08 & 520.92$\pm$358.4 & 80 \\
    
    &\begin{tabular}{@{}c@{}}Q-Learning \\ 1500 iterations\end{tabular}& 47.09$\pm$1.45 & 2.68$\pm$1.1 & 0.63 & \textbf{1.06$\pm$0.66} & 646.85$\pm$335.76 & 100 \\
    
    &\begin{tabular}{@{}c@{}}Q-Learning \\ Dynamic iterations\end{tabular}& \textbf{47.12$\pm$1.42} & \textbf{2.29$\pm$1.14} & \textbf{0.31} & \textbf{1.22$\pm$1.12} & \textbf{451.17$\pm$386.7} & 100 \\
\noalign{\smallskip}\hline
\end{tabular}
}
\end{table*}

In the indoor environment, the smallest distance and standard deviation were from A*, which tends to show the shortest possible path~\cite{toma2021pathbench}. On the other hand, RRT and Q-Learning with 300, 500, and dynamic iterations achieved results very close to A*. Also, the lower standard deviation shows the reliability of generating a similar trajectory for replanning. In addition to A*, Q-Learning with 150 iterations presented the best standard deviation. However, its trajectory is relatively longer.

Q-Learning with dynamic iterations also returned to the trajectory with the lowest average time, followed by A*. Moreover, the best time was for Q-Learning with dynamic iterations, probably when no obstacles were mapped. Q-Learning can quickly return trajectories when a replanning is close to the objective node.

Memory usage is similar for all algorithms except for PSO, which uses much more memory than other techniques, as this algorithm needs to save the values of several populations. The PSO processing was the smallest, as the PSO tries to find the shortest trajectory by optimizing a straight line between the start and end nodes. A* has the highest computational cost as it needs to calculate the value of most nodes in the scenario.

All algorithms present maximum completeness except for RRT, PSO, and Q-Learning, with 150 iterations. The RRT algorithm is a sampling-based technique and, thus, can not guarantee that a trajectory is always found. PSO, as it cannot continually continuously optimize the initial straight line to generate a collision-free trajectory. Furthermore, Q-Learning with 150 iterations, as this number of iterations is below what is necessary for this environment, so it was impossible to find a path without collision in many cases.

A* returned good time and distance results, but the computational cost is high. The RRT also showed promising results, but as its completeness was 65\%, it is unreliable to use it for online flights. The PSO has a low computational cost, but the trajectory size is large, and the completeness is low. Q-Learnings above 300 iterations showed completeness of 100\%, and dynamic iterations managed to return to the shortest trajectory in less time and with lower computational cost.

The smallest distance and standard deviation in the outdoor environment is A*. Regardless of the number of iterations, Q-Learning obtained the results closest to A*. All algorithms have a low standard deviation except for the PSO, which has the most extended trajectory for the forest environment.

RRT returned the lowest time average. RRT tends to be faster in forest environments as it explores more the paths.
After RRT, the shortest times were for A* and Q-Learning with dynamic iterations, respectively. The best times were from A* and RRT. The dynamic Q-Learning presents the best time among the simulations with Q-Learning.

The memory usage is lower for Q-Learning techniques with a higher number of iterations, as they learn to move around in the environment in the training phase, so in the test phase, they only follow the shortest route. Again, in computational cost, A* had the highest cost because it needed more analysis in the scenario. On the other hand, PSO and RRT had the lowest computational cost. It occurs in PSO for the same reasons observed in indoor environments. On the other hand, in the case of RRT, it occurs due to the presence of numerous open spaces leading toward the goal, facilitating the generation of tree structures.

In this case, the maximum completeness algorithms were A*, Q-Learning with 1500, and dynamic iterations. The RRT and PSO did not obtain maximum completeness in the indoor environment and are also valid for the outdoor environment. Q-Learning with 800 iterations is below the average needed to complete the trajectory in this environment, as shown in Figure~\ref{fig:reward}. Hence, it had problems finding a trajectory in some simulations. The same happened with Q-Learning 1000 iterations, even using the recommended average to find a trajectory. Q-Learning, with 1500 iterations, achieved its maximum completeness for being above what is necessary to determine a trajectory. Q-Learning with dynamic iterations achieved it because it only stops training when it has learned to move around in the environment.

A* returned the best results except for computational cost, which had the highest among all algorithms. The RRT returned the trajectories in the shortest time. However, it obtained one of the most extended distances, with 71\% completeness. Finally, the Q-Learning with dynamic iterations presented the best results among the simulations with Q-Learning. Besides presenting the lowest computational cost among the techniques obtained 100\% completeness and a trajectory size near A*.

\begin{figure*}
    \centering
    \begin{subfigure}[b]{0.475\textwidth}
        \centering
        \includegraphics[width=\textwidth]{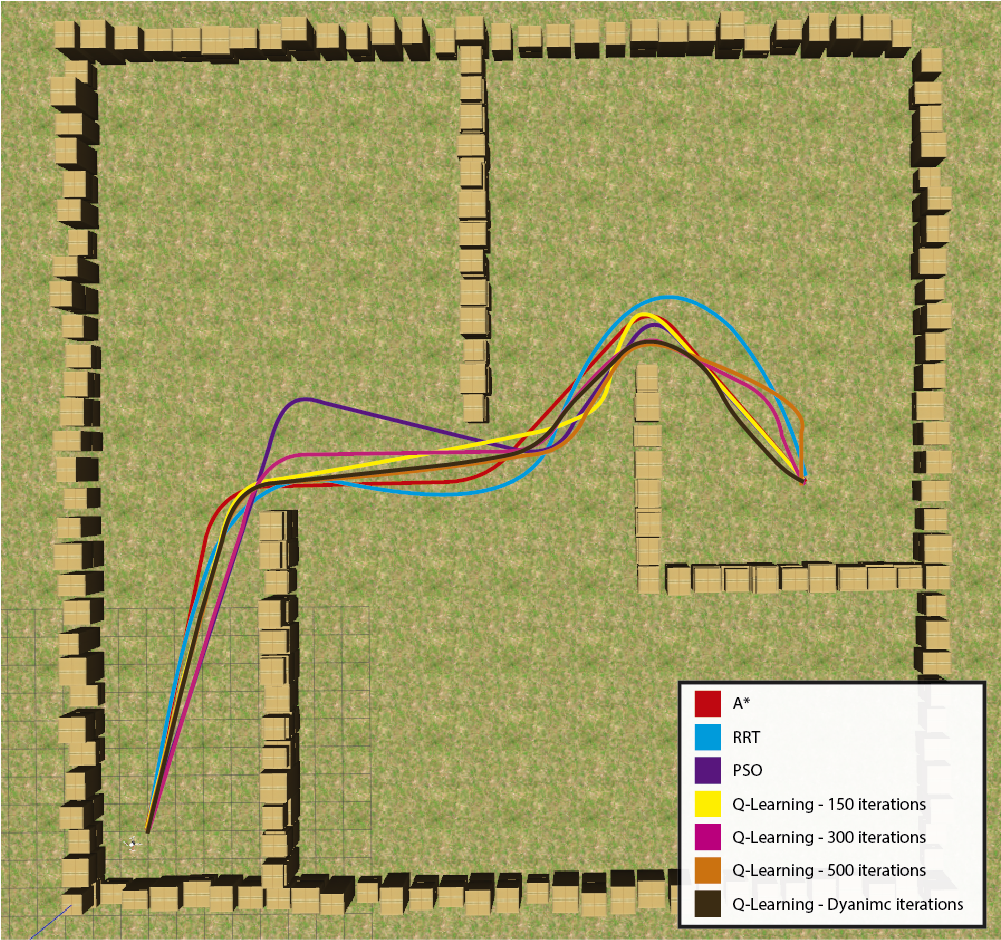}
        \caption{}
    \end{subfigure}
    \hfill
    \begin{subfigure}[b]{0.475\textwidth}
        \centering
        \includegraphics[width=0.94\textwidth]{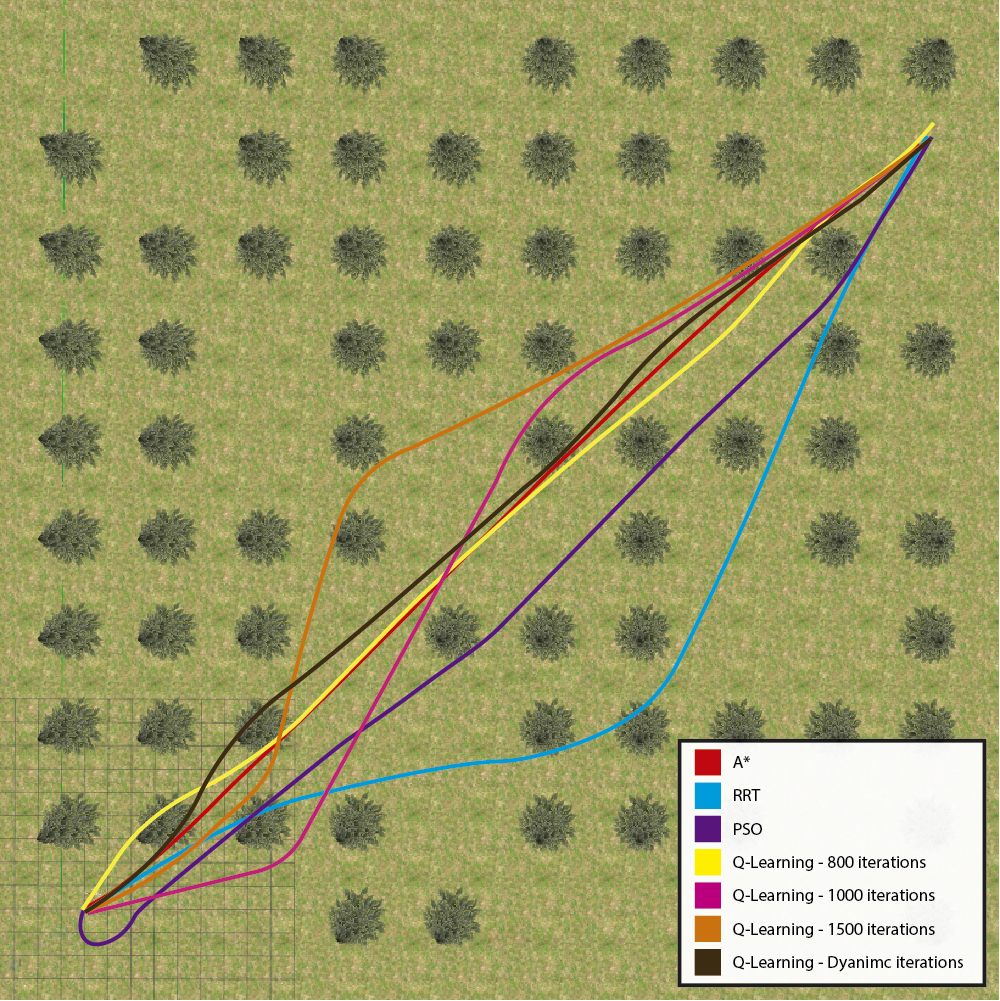}
        \caption{}
    \end{subfigure}
    \caption{Comparison of paths between the best results of the 100 iterations of the A*, RRT, PSO, Q-Learning techniques.}
    \label{fig:comparison}       
\end{figure*}

In Figure~\ref{fig:comparison} the trajectories of all these algorithms are compared. This comparison serves for the dual purpose of assessing the proximity of the Q-Learning trajectory to the approaches in the literature and evaluating their suitability for real UAV utilization.

All the trajectories in the indoor environment are smoothed and obey the constraints of the UAVs. However, the trajectories generated by the PSO and the RRT follow a noticeably longer path than the others. All other trajectories reached the goal along similar paths.

The trajectories of the outdoor environment are also good to be followed. Except for the RRT and Q-Learning with 1000 and 1500 iterations, returned trajectories were visibly more extensive than the others, as they were not just a straight line deviating from the trees. They sought a safer trajectory around them.

Q-Learning with dynamic iterations yielded superior results in terms of distance, time, computational cost, and completeness compared to the original Q-Learning with any fixed value of iterations. It was also better than PSO on all metrics except computational cost. It surpassed the RRT in all metrics except for the outdoor environment's response time and computational cost. Despite not having returned trajectories shorter than A*, it returned trajectories in adequate time. It had the best indoor scenario, besides having a much lower computational cost and the same completeness as A*.

Within 3D environments, the generated paths follow the same pattern due to the Q-Learning policy that focuses on finding the shortest path with no collisions. Moreover, our optimization efforts are intricately aligned with the specific constraints inherent to UAVs. This approach is geared towards not only maintaining uniformity in trajectory patterns but also ensuring that the resultant 3D trajectories are as feasible as the 2D trajectories.

\section{Applications in Real World}
\label{sec:real}
This section explores the potential applications of dynamic Q-Learning in real-world scenarios, considering its advantages over classical and meta-heuristic techniques. Classical techniques are commonly employed to generate point-to-point trajectories or serve as guidance paths for UAVs. In static environments, exact classical algorithms are utilized with offline planning methods due to their high computational cost. However, these algorithms prove unreliable in real-time operations, particularly in unknown or expansive environments. As a result, approximated classical algorithms are often employed, offering feasible trajectories within reasonable timeframes but built with random samples that can result in challenging UAV maneuvers. Moreover, replanning with these techniques must be more reliable as they produce disparate trajectories without a systematic assessment of the optimal option for the environment, potentially introducing drastic changes to the flight path and increasing its complexity.

Meta-heuristic techniques are not extensively employed due to their time-consuming trajectory generation process, rendering them unsuitable for real-time applications. Additionally, the trajectories they generate may only be partially feasible, necessitating continuous trajectory optimization when employing these algorithms. Although meta-heuristic trajectories are typically smooth and straightforward, enabling easier UAV navigation, the drawbacks associated with these approaches limit their reliability in real-world flights.

In contrast, machine learning techniques, such as Q-Learning, offer distinct advantages. Notably, the computational cost remains constant regardless of the environmental complexity, making them well-suited for real-world applications. Q-Learning algorithms learn to navigate the environment by adhering to a predefined policy that penalizes collisions with obstacles. Through iterative iterations, the algorithm progressively learns to reach the goal efficiently while avoiding obstacles. This learned policy is saved in the Q-Table, enabling reliable replanning in real-time.

Moreover, Q-Learning trajectories exhibit similarities to those generated by meta-heuristic techniques, characterized by smooth and straightforward paths that allow for increased speed when necessary. However, the main drawback of Q-Learning lies in the time required to learn the environment, as a short training time may result in unreliable trajectories. In contrast, an excessively long training time can be impractical for replanning. Considering it, we proposed the dynamic Q-Learning to determine the optimal number of iterations for each flight, optimizing the training time while ensuring reliable trajectory generation.

With our optimization, trajectory replanning in both indoor and outdoor environments is achieved within a remarkable 1-second timeframe, as exemplified in Section~\ref{sec:simulations}. This capability allows the drone to detect obstacles at a distance equivalent to its one-second travel distance. For instance, assuming a drone speed of 4 m/s, the system aims to identify obstacles at a distance of 4 to 5 meters, incorporating an error margin. Notably, contemporary cameras typically detect obstacles within a range of 10 to 15 meters, while lidars extend this capability up to 50 meters. This affords the UAV a secure timeframe for real-time trajectory replanning. Moreover, it is pertinent to highlight that a drone flying at 4 m/s is considered a high-speed autonomous flight, with conventional speeds ranging between 1 to 2 m/s. Therefore, our methodology ensures ample time for algorithmic validation and real-time replanning, emphasizing the safety and reliability of the proposed approach.

Dynamic Q-Learning can benefit various missions, including:

\begin{enumerate}
    \item Monitoring: Dynamic Q-Learning enables real-time trajectory generation in complex environments, facilitating interaction with the environment and adaptability to new monitoring goals. This capability allows for seamless monitoring of designated areas and the ability to replan trajectories for newly identified regions dynamically.
    \item Delivery and agriculture: In time-sensitive missions like delivery or agricultural tasks, efficient flights are crucial to optimize resource utilization, such as battery life, and to achieve rapid goal attainment. By leveraging dynamic Q-Learning, UAVs can follow smooth, straightforward trajectories, enabling high-speed flights. Additionally, faster replanning with the dynamic approach enhances reliability during high-speed operations.
    \item Complex environments (e.g., forests and mines): These environments often involve multiple tasks within the same area. Dynamic Q-Learning proves advantageous in such scenarios as UAVs can learn to navigate the environment irrespective of the starting and goal nodes. Consequently, reliable real-time replanning becomes possible, even in the environment's presence of changes or unpredictability.
\end{enumerate}

By applying dynamic Q-Learning to various missions, UAV operations can benefit from optimized trajectory planning, improved adaptability to environmental changes, and enhanced efficiency in real-world applications.

\section{Conclusion}
\label{sec:conclusion}
In this paper, we have presented an online path planning approach based on Reinforcement Learning specifically designed for unknown and complex environments. Our algorithm incrementally builds a map through exploration, enabling effective path planning in unknown environments. Extensive experiments were conducted in both indoor and outdoor unstructured and unknown environments, focusing on evaluating various metrics including path length, time, memory usage, CPU utilization, and completeness.

The experimental results revealed that the widely used A* algorithm consistently provided the best distance and time results in both scenarios. However, it came at a high computational cost. In contrast, our proposed Q-Learning planner with dynamic iterations showcased remarkable performance with favorable metrics after A*, while maintaining low computational overhead and ensuring maximum completeness.

Overall, this research contributes by introducing a novel approach to online path planning, tailored for challenging and unknown environments. The extensive experiments and comparative analysis provide valuable insights into the trade-offs between performance, computational efficiency, and completeness. These findings underscore the potential of our proposed algorithm as a viable solution for real-world applications that require efficient and reliable path planning in unknown and complex environments.

\section{Acknowledgements}
The authors would like to thank CAPES and CNPq for their financial support. Also, the authors would like to be thankful for the partnership with the University of Sao Paulo, the University of Sydney, and the Czech Technical University in Prague.

\end{document}